\RequirePackage{amsthm}

\documentclass[sn-mathphys,Numbered]{sn-jnl}

\usepackage{graphicx}%
\usepackage{multirow}%
\usepackage{amsmath,amssymb,amsfonts}%
\usepackage{amsthm}%
\usepackage{mathrsfs}%
\usepackage[title]{appendix}%
\usepackage{xcolor}%
\usepackage{textcomp}%
\usepackage{manyfoot}%
\usepackage{booktabs}%
\usepackage{algorithm}%
\usepackage{algorithmicx}%
\usepackage{algpseudocode}%
\usepackage{listings}%
\usepackage{siunitx}

\theoremstyle{thmstyleone}%
\theoremstyle{thmstyletwo}%

\theoremstyle{thmstylethree}%

\begin{document}

\title[Article Title]{Circuit-centric Genetic Algorithm (CGA) for Analog and Radio-Frequency Circuit Optimization}


\author[1]{\fnm{Yeonjun} \sur{Lee}}\email{yeonjun3333@gmail.com}
\equalcont{These authors contributed equally to this work.}

\author[1]{\fnm{Mingi} \sur{Kwon}}\email{mingi001025@gmail.com}
\equalcont{These authors contributed equally to this work.}

\author*[1]{\fnm{Ickhyun} \sur{Song}}\email{isong@hanyang.ac.kr}

\affil[1]{\orgdiv{Department of Electronic Engineering}, \orgname{Hanyang University}, \orgaddress{\street{222 Wangsimni-ro}, \city{Seongdong-gu}, \postcode{04763}, \state{Seoul}, \country{Republic of Korea}}}


\abstract{This paper presents an automated method for optimizing parameters in analog/high-frequency circuits, aiming to maximize performance parameters of a radio-frequency (RF) receiver. The design target includes a reduction of power consumption and noise figure and an increase in conversion gain. This study investigates the use of an artificial algorithm for the optimization of a receiver, illustrating how to fulfill the performance parameters with diverse circuit parameters. To overcome issues observed in the traditional Genetic Algorithm (GA), the concept of the Circuit-centric Genetic Algorithm (CGA) is proposed as a viable approach. The new method adopts an inference process that is simpler and computationally more efficient than the existing deep learning models. In addition, CGA offers significant advantages over manual design of finding optimal points and the conventional GA, mitigating the designer's workload while searching for superior optimum points.
}




\maketitle
\newpage
\section{Introduction}\label{sec1}

Artificial intelligence (AI) has made significant advancements in various fields and it has been widely applied across various domains. Continuous semiconductor scaling enhances hardware performance, while improvements in the quality of vast input data enable AI to surpass human capabilities in numerous areas. In this regard, the circuit design process is also under consideration for AI applications, given its need to deal with complex interactions between mathematical calculations and various components. The optimization algorithms proposed so far, however, are computationally intensive and resource-hungry, resulting in prolonged design times, reduced design process efficiency, and increased economic costs due to excessive power consumption. \cite{r1}

In the literature, there have been several approaches to circuit design and optimization. Many researchers have focused on circuit creation based on expert knowledge \cite{r2}, \cite{r3}. However, making a good technical decision within a vast design space that encompasses all possible circuits is challenging. Whereas several nature-inspired algorithms such as PSO (particle swarm optimization) and ACO (ant colony optimization) have shown promise in solving problems, however, they still face difficulties in terms of accuracy and computation time for multivariate and multimodal problems.\cite{r4}

Genetic Algorithms (GAs) are recognized for their efficiency in conducting comprehensive global searches. They excel in maintaining a population of diverse solutions, enabling the exploration of extensive search spaces. These attributes render GAs particularly well-suited for optimizing circuit designs, a domain characterized by intricate, multimodal, and non-convex fitness landscapes that challenge the identification of global optima. Furthermore, GAs exhibit a high degree of adaptability, making them versatile across various problem domains and solution representations. Their applicability extends seamlessly to both discrete and continuous optimization problems. GAs empower users to finely tailor their choice of encoding and genetic operators to precisely align with the specific nuances of the given problem. Additionally, the intrinsic ability of GAs to uphold genetic diversity within their populations is invaluable. Mechanisms like mutation and recombination facilitate this diversity maintenance \cite{r5}. Such inherent diversity plays a pivotal role in guiding GAs away from local optima, allowing for a more thorough exploration of the search space. GAs offer a powerful and adaptable approach for circuit design optimization, particularly in scenarios with complex fitness landscapes, setting them apart from PSO and ACO.\cite{r6} practice.

High-frequency (RF) circuits are composed of various parameters, including noise, voltage gain, input-output matching, stability, power consumption, and noise figure, all of which must be satisfied. RF circuit design is characterized by heightened sensitivity to parameter variations, where even slight changes can significantly impact performance. This heightened sensitivity extends to both analog and high-frequency characteristics, demanding a comprehensive consideration of parameters in both domains. The intricate interplay of analog and high-frequency traits necessitates meticulous attention to factors such as impedance matching, signal integrity, and power handling. Furthermore, the influence of noise becomes substantially more pronounced in RF circuits. The impact of noise on signal quality and system performance is magnified, requiring sophisticated noise mitigation strategies and careful design considerations. In the delicate balance of optimizing parameters, RF circuit designers must navigate the challenges posed by the increased susceptibility to variations and the pervasive influence of noise, making the design process inherently more intricate and demanding.

 RF circuits and transceivers involve numerous trade-offs, especially as circuits become more complex \cite{r7}. Quantitative analysis of the impact of parameters on the final performance and design variables becomes limited as circuits become more complex, and manually finding optimal points for each parameter as analog circuit design parameters increase heavily depends on the designer's experience and heuristic approaches. Trade-off relationships between specific performance metrics and others can complicate the optimization process. For example, lowering the noise of a front-end amplifier may require higher power consumption or sacrifice linearity \cite{r7}.

To address these challenges, this paper introduces a modified GA termed circuit-centric genetic algorithm (CGA) to conduct receiver (Rx) optimization, offering a simple yet effective approach. An algorithm suitable for transistor-level circuit design transfer was developed by modifying the mutation method and generational transfer method in traditional GA. Research is currently underway to enhance the efficiency of circuit design engineers using Python-based automation to control Cadence tools and other design automation techniques \cite{r18}. This study was also written in Python and utilized the free commercial tool LTspice to automatically optimize circuit performance. The optimization results were evaluated using the figure of merit (FoM), which includes key performance metrics for the receiver, such as gain, power consumption, and noise figure.

The proposed circuit design in this study demonstrated a 30\% improvement in FoM compared to existing circuits, validating the efficiency of circuit design using genetic algorithms and showcasing its applicability to various analog circuit designs.

\section{Receiver Design and Figure of Merit} \label{sec2}
One of the key modules in general RF systems is an RF receiver (Rx), which essentially performs detection, amplification, and frequency translation of the incoming RF inputs, resulting in an intermediate frequency (IF) signal for subsequent data processing \cite{r8}. The block diagram of a typical Rx is shown in Fig. 1. An antenna is used to receive wireless signals at a specific frequency. It captures radio waves through the air, converts them into electrical signals, and transmits them to the receiver. As the first gain stage of an Rx an LNA is used to amplify a very weak signal from the antenna with minimal addition of noise to the input. An ideal LNA should provide a high gain and a low noise figure (NF) \cite{r9}. After the LNA stage, a down-conversion mixer conducts frequency-mixing operation, using a local oscillator (LO) signal from a voltage-controlled oscillator (VCO) and generating a voltage output at intermediate frequency (IF). LO is a local oscillator commonly used in receivers. It generates a specific frequency to select the desired frequency without interference from other signals. The VCO, responsive to changes in voltage, allows for dynamic control of the LO's frequency. Since a mixer may generate relatively high noise, it usually is placed after the LNA for noise minimization of the entire Rx chain \cite{r7}. Additionally, placing an IF Filter after the down conversion mixer is crucial to isolate the desired signal from multiple frequency components and enhance the receiver's selectivity for effective signal extraction \cite{r10}.

\begin{figure}[h]%
\centering
\includegraphics[width=0.9\textwidth]{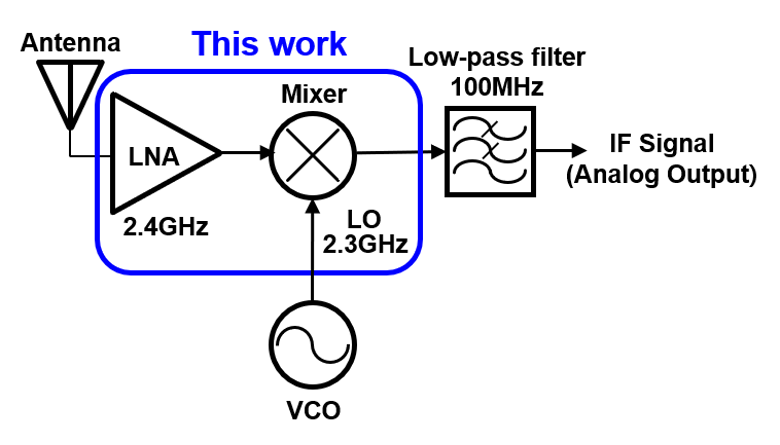}
\caption{System configuration of an RF receiver (Rx) in which an antenna, an LNA, a down-conversion mixer, a voltage-controlled oscillator (VCO), and a low-pass filter are included. The frequencies of the input, the local oscillator (LO) signal, and the output are 2.4 GHz, 2.4 GHz, and 100 MHz, respectively.
}\label{fig1}
\end{figure}

The overall configuration of the Rx is based on the example in  Razavi's 'RF Microelectronics 2nd edition' \cite{r7} and the combination of the LNA and the mixer will be explored throughout this study designed for operation in the 2.4GHz band. The circuit includes an LNA for noise reduction and increased gain, with a down-conversion mixer connected at the back of the LNA. The entire receiver is configured for down conversion to a 100MHz IF frequency using a 2.3GHz Local Oscillator. The power supply voltage is set at 1.2V.  Single Balanced Mixer (SBM) is selected for its simplicity and versatile performance, making it suitable for general-purpose use compared to the more complex Double Balanced Mixer (DBM). The LNA and SBM are designed using \SI{0.18}{\micro\metre} CMOS Technology \cite{r10}. The circuit was designed and simulated using built-in components in LTspice, and LTspice was controlled using Python for automation.

\begin{figure}[h]%
\centering
\includegraphics[width=0.9\textwidth]{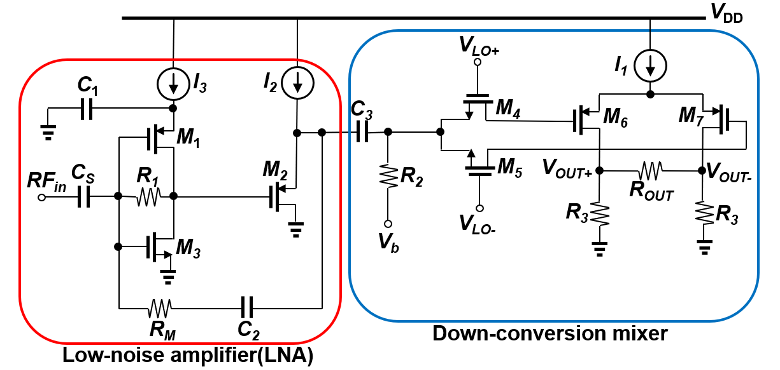}
\caption{Full schematic of RF receiver  including an LNA and a single-balanced down -conversion mixer.
}\label{fig2}
\end{figure}

The designed circuit as shown in Fig. 2. represents the block before C3 as an LNA, followed by a mixer in the next part. PMOS M1 serves as both a load and an amplifying device, yielding a lower noise figure than if the load is passive. Current source I1 defines the bias of M1 and M2, and C1 creates an ac ground \cite{r7}. NMOS M2 is connected to act as a buffer for the LNA. Setting R1 with a large value facilitates the establishment of an appropriate DC level at the gates of M1 and M3, thereby enabling a higher voltage gain. The mixer is designed by increasing the PMOS sizes of M6 and M7 to minimize flicker's noise. Vb defines the gate bias voltage for the differential pair, maintaining a level of 0.2 V \cite{r7}.

The LNA plays a role in amplifying the input signal and minimizing noise, while the mixer performs downconversion for data processing at intermediate frequencies. Strongly coupling these two components enhances the overall circuit performance, and it allows finding optimal solutions in cases where certain trade-offs occur. Therefore, optimizing the LNA and mixer as a unified block can improve the efficiency and performance of the entire receiver system.

Although there are many performance factors in LNAs, we concentrate on only three important parameters such as voltage gain, noise figure and power dissipation \cite{r11}. 

Considering that in a cascaded circuit, the noise figure is mainly affected by the value of the first block, the noise figure for the entire receiver was approximated by solely measuring the noise figure of the LNA. (The LNA gain obtained from the actual results is greater than 10dB, so taking the mixer's noise figure into account did not significantly change the figures.) The overall noise figure of the cascaded circuit was calculated by referencing the Friis equation as follows:
\begin{equation}
F_{\text{total}} = F_1 + \frac{F_2 - 1}{G_1} + \frac{F_3 - 1}{G_1G_2} + \frac{F_4 - 1}{G_1G_2G_3} + \ldots + \frac{F_{n-1} - 1}{G_1G_2G_3\ldots G_{n-1}}
\end{equation}

For the measurement of downconversion gain, the large signal model was applied to calculate the maximum differential output voltage of the receiver's output relative to the maximum RF voltage input,set at an amplitude of 0.3. Power consumption was determined by averaging the power consumed by all circuit components during the entire 50ns operational period.

The entire algorithm progressed through simulations aimed at maximizing the FoM by calculating three parameters. However, it was necessary to prevent the FoM from excessively increasing in the undesired direction of the circuit's specifications. For instance,  where the noise figure exceeds 5 but the down conversion gain significantly compensates for it, the value of the noise figure was replaced with 10,000 to hinder an anomalous increase in the FoM. This method was taken to design the circuit within the desired specification range.

\begin{equation}
FoM = \frac{Conversion gain [dB]}{Noise figure [dB] \cdot Power consumption [W]}
\end{equation}

\section{Genetic Algorithm Adaptation for Analog and Radio-frequency Circuit}\label{sec3}

\subsection{Genetic Algorithm}\label{sec3_1}
A Genetic Algorithm (GA) is a heuristic search algorithm used to solve optimization problems. It is based on the fundamental concept that "all organisms survive by adapting to various environments," which was proposed by Charles Darwin's theory of natural selection. Genetic Algorithms were first introduced by John Holland in the 1960s and have since been applied as a paradigm of natural evolution \cite{r16}. They are based on concepts such as genes, mutations, natural selection, and crossover.

Utilizing these concepts, GAs form a new candidate solution set and iteratively find the optimal solution over multiple generations. The candidate solutions within each generation's solution set are referred to as chromosomes or individuals \cite{r17}.

\subsection{Advantages of Genetic Algorithm}\label{sec3_2}
Genetic Algorithms do not require constraints such as continuity or differentiability of the objective function, unlike conventional methods. This is because designers determine the fitness function in a way that promotes an increase in the desired direction, and the algorithm solely computes this function's values for optimization. Therefore, Genetic Algorithms offer the advantage of being relatively easy to apply to real-world systems without the need for additional information beyond fitness evaluation.

\subsection{Basic Principles of Genetic Algorithm} \label{sec3_3}
\begin{figure}[h]%
\centering
\includegraphics[width=0.9\textwidth]{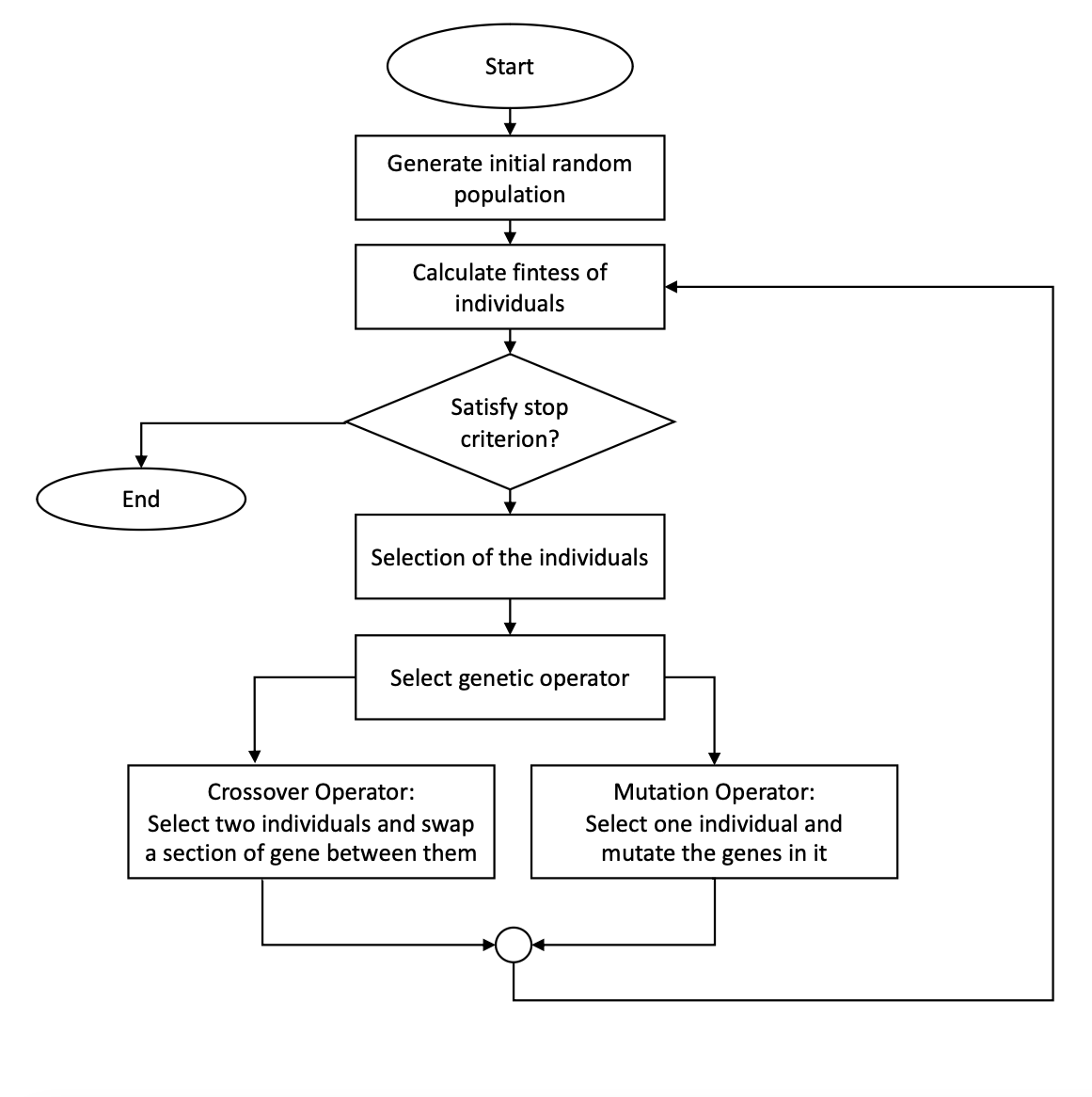}
\caption{Genetic Algorithm flowchart\cite{r14}}
\label{fig3}
\end{figure}

Genetic Algorithms are based on concepts such as genes, natural selection, crossover, and mutation. Natural selection is the process of determining which individuals within the current generation will be passed on to the next generation. Typically, individuals with higher fitness are selected. \cite{r12} The fitness function measures how good each individual's solution is, and in this study, FoM was used. FoM is calculated as the ratio of Gain to the product of Power Consumption and Noise Figure, where higher FoM values represent superior performance.

Crossover, on the other hand, involves exchanging information between selected parent individuals to create new offspring individuals. Crossover can be performed in various ways, depending on the characteristics of the Genetic Algorithm. \cite{r14}

Mutation introduces diversity by randomly altering an individual's genes. It occasionally introduces new information and makes the optimization process more exploratory.

In this paper, the algorithm designed is more suitable for circuit design automation by omitting the crossover process and enhancing the mutation process. First, the drawbacks of using the unaltered Genetic Algorithm are discussed, and then the improved algorithm is described to highlight its potential for circuit enhancements.

\subsection{ The Challenges of Applying Genetic Algorithms in Circuit Design}\label{3_4}
\begin{figure}[h]%
\centering
\includegraphics[width=0.9\textwidth]{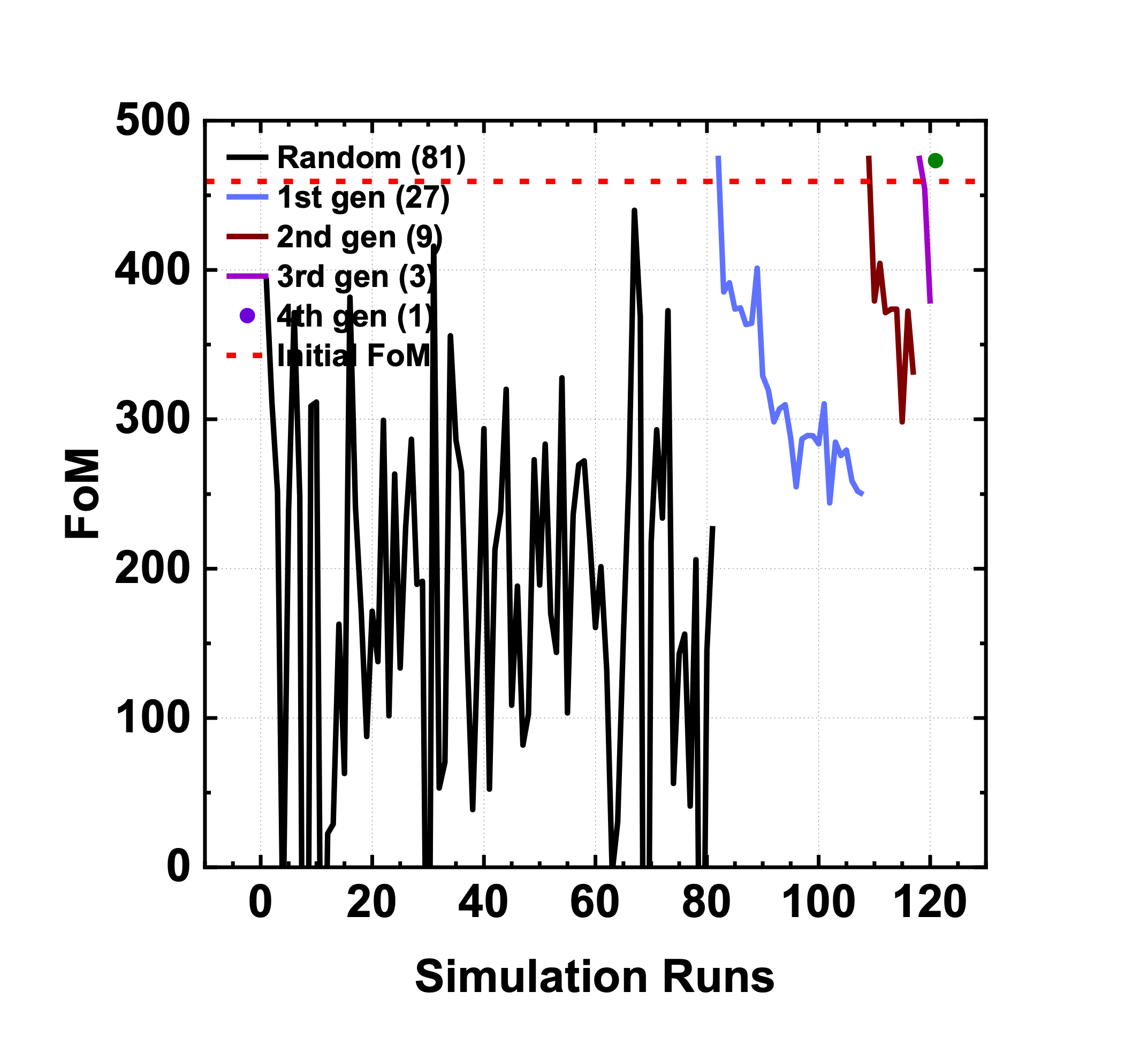}
\caption{FoM of circuit designed with original GA algorithm}\label{fig4}
\end{figure}

The graph above represents the FoM graph computed using the Genetic Algorithm (GA), which effectively illustrates why GA may not be suitable for circuit design.

Optimization was carried out over a total of five generations. In the initial generation, 81 individuals were randomly generated. The FoM values obtained were then sorted in descending order, and the top 1/3 of individuals, which corresponds to the top 27 individuals, were selected. Subsequently, a crossover operation was applied to these selected individuals. From the graph, it is evident that in the first generation, the individual with the highest FoM is the 67th individual, with a FoM of 439.92.

One randomly selected component among the twenty components of this individual was specified. For this specified component, a new random value was generated within the predefined range for that component. FoM was recalculated for the individual after this mutation operation. In the case of the 67th individual, after the mutation operation, its FoM increased to 476.58.
\begin{figure}[h]%
\centering
\includegraphics[width=0.9\textwidth]{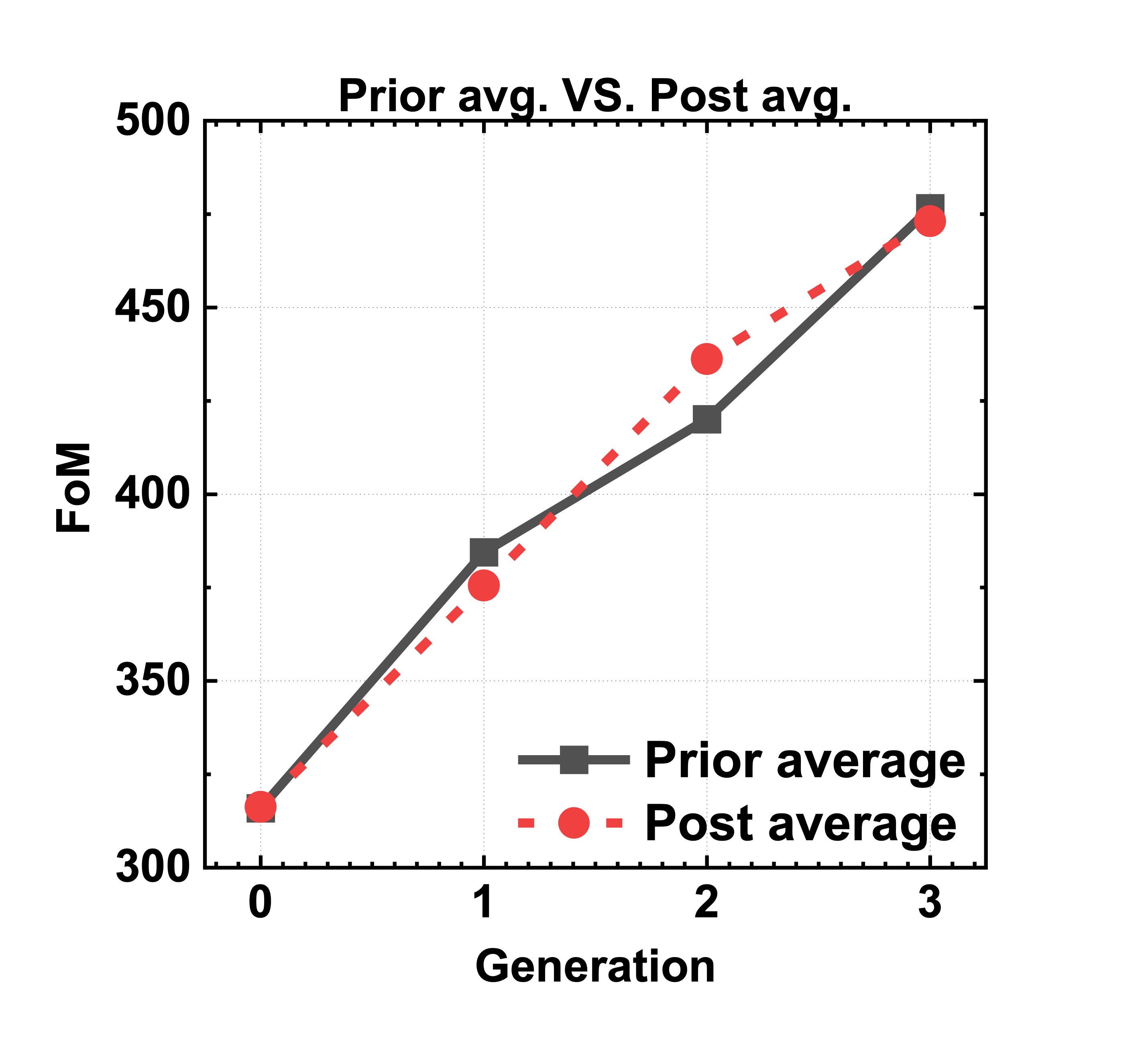}
\caption{System configuration of an RF receiver (Rx) in which an antenna, an LNA, a down-conversion mixer, a voltage-controlled oscillator (VCO), and a low-pass filter are included. The frequencies of the input, the local oscillator (LO) signal, and the output are 2.4 GHz, 2.4 GHz, and 100 MHz, respectively.
}\label{fig5}
\end{figure}

The graph above highlights the inherent issues with applying the genetic algorithm directly to this research. It becomes evident that Post consistently does not exhibit higher FoM values compared to Prior. The average FoM of the top 27 individuals in the initial generation is 315.94, but the average FoM for the second generation, which undergoes mutation operations on these initial individuals, only slightly increases to 316.28, indicating insufficient performance improvement.

Furthermore, the average FoM of the top 9 individuals in the second generation is 384.45, yet the average FoM of the third generation, which comprises the results of the mutation operations on these top-performing individuals, decreases to 375.56. This contradicts the fundamental principle of genetic algorithms, where over generations, only superior individuals should survive and evolve.

This problem arises because individuals with initially high FoM values tend to lose their distinctive characteristics through information exchange with other individuals during the crossover process. Therefore, the mutation operations via genetic algorithms do not effectively facilitate the clear learning and improvement of FoM.

\subsection{CGA : Circuit-Centric Genetic Algorithm}\label{sec3_5}
This paper addresses the problem of finding optimal component values for LNA (Low Noise Amplifier) and Mixer circuits. Genetic Algorithms (GAs) typically generate offspring generations through the crossover of superior individuals. However, when applying GA to this research, it encountered issues of randomness, such as a significant drop in the overall FoM due to the crossover of circuits where Noise Figure values dominated, despite the component values for Gain being superior.

In response to these challenges, a new approach called Circuit-centric Genetic Algorithm (CGA) was devised and adopted.

\begin{algorithm}
\caption{Genetic Algorithm for Electronic Circuit Design}
\begin{algorithmic}
\State Initialize parameters:
\State \hspace{1cm} $n_{\text{gen}}$, $\text{pop\_size}$, $\text{target\_FOM}$, $\text{max\_simulations}$
\State Initialize $\text{best\_FOM}$, $\text{best\_individual}$, $\text{simulations\_run}$ to 0
\State Create an initial population of individuals:
\State \hspace{1cm} Each individual represents a set of circuit parameters
\State \hspace{1cm} Randomly generate these parameters within specified ranges
\While{$\text{simulations\_run} < \text{max\_simulations}$ \textbf{and} $\text{best\_FOM} < \text{target\_FOM}$}
    \State Evaluate each individual in the population:
    \ForAll{individuals}
        \State Circuit modification
        \State Run LTspice simulations to obtain:
        \State \hspace{1cm} Gain
        \State \hspace{1cm} Power consumption
        \State \hspace{1cm} Noise figure
        \State Calculate the Figure of Merit (FOM)
    \EndFor
    \State Sort the evaluated population based on FOM (highest to lowest)
    \State Select the best-performing individual
    \ForAll{selected individuals}
        \State Mutatation
        \State Run LTspice 
        \State If the FOM improves, update the individual;
    \EndFor
    \State Check termination conditions:
    \If{$\text{best\_FOM} \geq \text{target\_FOM}$ \textbf{or} $\text{simulations\_run} \geq \text{max\_simulations}$}
        \State Exit loop
    \EndIf
    \State Update the population with the newly created individuals
\EndWhile
\end{algorithmic}
\end{algorithm}

\subsection{CGA operation}\label{sec3_6}
The provided pseudo code outlines the modified CGA algorithm. In the initial generation, 30 individuals are randomly generated. Among these generated individuals, the one with the best performance, meaning the highest FoM, is selected. For this chosen individual, the mutation algorithm is applied a total of 20 times. The reason for performing this operation 20 times is to apply a mutation operation once for each component variable.

From these newly modified twenty individuals, the one that exhibits the highest FoM is reassigned. The same process is then repeated for this re-assigned individual. By employing this adapted algorithm, the system is designed to continuously improve the component values of individuals that demonstrate superior performance.

\subsection{CGA’s distinct feature compared to GA}\label{sec3_7}
The key distinctions between CGA and the traditional GA lie in the absence of a crossover process in CGA and the continuous application of mutation operations exclusively to a single individual. In traditional GA optimization, a certain proportion of individuals from the current generation typically engage in crossover. However, in CGA, instead of crossover a fraction of individuals from the generation, the chosen approach involves performing mutation operations for the best FoM individual for each component variable, and this is repeated twenty times.

This fundamental difference between GA and CGA addresses the issue at hand. The improved algorithm boldly focuses on a single individual with the highest FoM, bypassing others, and applies mutation operations for all twenty component variables. This significantly increases the possibility of improving FoM. While it might be conceivable to perform mutation operations for multiple individuals with top FoM values, the decision to focus on a single top-performing individual is made to preserve the simplicity inherent in genetic algorithms and reduce computational complexity.

This approach ensures that FoM can continuously improve over generations by reintroducing randomness while preserving parameters that maintain a favorable FoM, which sets it apart from the traditional GA.

\section{Optimization Results and Discussion}\label{sec4}
\subsection{Results}\label{subsec4}

\begin{figure}[h]%
\centering
\includegraphics[width=0.9\textwidth]{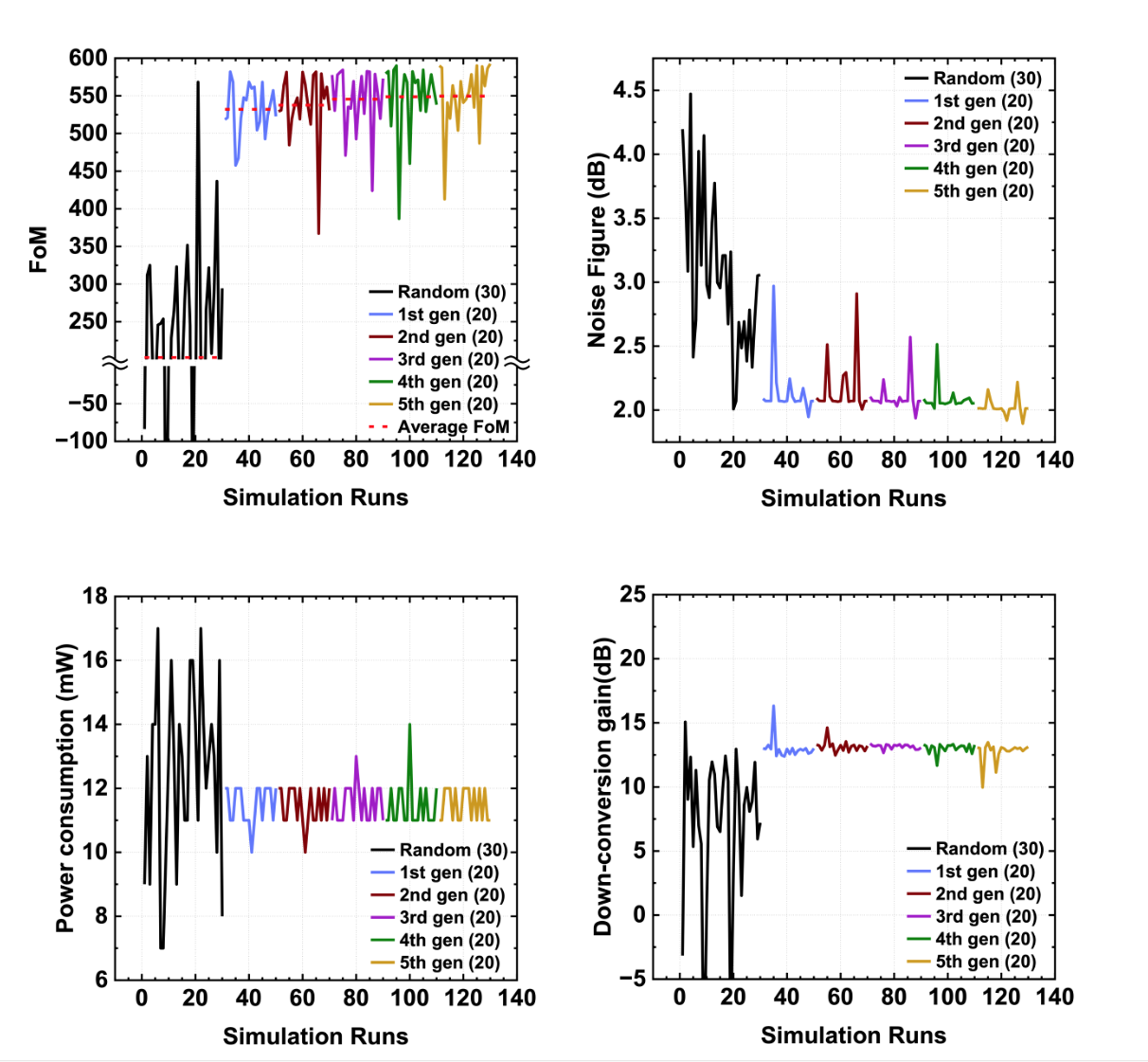}
\caption{FoM steadily rising with its parameters gain, power consumption and noise figure
}\label{fig6}
\end{figure}

When examining the FoM graphs, it becomes evident that unlike traditional GA, which did not consistently show an upward trend with each generation, CGA displays a clear upward trajectory. This increase is not merely reliant on the randomness of the initial generation but is confirmed to be a result of the principles of the optimization algorithm. Observing the graphs of individual components contributing to FoM, it is notable that Gain increases, while power consumption and noise figure decrease. This illustrates that FoM improvement is not solely driven by a single parameter but involves the simultaneous exploration of optimal values for all three parameters.

Notably, the significant improvement in the final FoM is predominantly driven by the lower noise figure, contributing to approximately a 43.5

This flexibility in CGA allows circuit designers to easily tailor the FoM formula or set threshold values for specific parameters based on their priority for improvement. This enables the straightforward adjustment of the FoM to focus on specific aspects of circuit performance, depending on the designer's objectives.

\begin{table}[h!]
\centering

\caption{Optimal circuit component variables drawn by the Circuit-centric Genetic Algorithm}

\begin{tabular} { |c|c|c|c|c| }
\hline
Parameter & Initial Value & Optimal Value & Range & Step size\\
\hline
R1 (\si{\ohm}) & 5,000 & 5,146 & 3,000 -- 6,000 & 1\\
R2 (\si{\ohm}) & 2,000 & 3,219 & 1,500 -- 4,000 & 1 \\
R3 (\si{\ohm}) & 200 & 897 & 1 -- 1,000 & 1 \\
Rout (\si{\ohm}) & 100,000 & 8,078 & 1,000 -- 100,000 & 1\\
Rm (\si{\ohm}) & 450 & 998 & 100 -- 1,000 & 1\\ 
C1 (\si{\pico\farad}) & 2 & 2.92 & 1.5 -- 3 & 0.01\\
C2 (\si{\pico\farad}) & 1 & 1.05 & 0.1 -- 5 & 0.01 \\
C3 (\si{\pico\farad}) & 0.5 & 0.86 & 0.1 -- 5 & 0.01 \\
Vb (V) & 0.2 & 0.11 & 0.1 -- 0.4 & 0.001\\
I1 (\si{\milli\ampere}) & 5 & 3.69 & 3 -- 7 & 0.01\\
I2 (\si{\milli\ampere}) & 2 & 3.33 & 1 -- 5 & 0.01\\
I3 (\si{\milli\ampere}) & 2 & 3.56 & 1 -- 5 & 0.01\\
VLO offset (V) & 0.5 & 0.66 & 0.3 -- 0.7 & 0.01 \\
VLO Amp (V) & 0.4 & 0.65 & 0.2 -- 1.0 & 0.01 \\
M1 width (\si{\micro\meter}) & 144 & 101 & 90 -- 180 & 1\\
M2 width (\si{\micro\meter}) & 14.4 & 13.57 & 9 -- 18 & 0.01\\
M3 width (\si{\micro\meter}) & 5.4 & 4.04 & 1.8 -- 9 & 0.01\\
M4 width (\si{\micro\meter}) & 0.9 & 1.63 & 0.54 -- 1.8 & 0.01\\
M5 width (\si{\micro\meter}) & 0.9 & 1.63 & 0.54 -- 1.8 & 0.01\\
M6 width (\si{\micro\meter}) & 15 & 16.81 & 9 -- 18 & 0.01\\
M7 width (\si{\micro\meter}) & 15 & 16.81 & 9 -- 18 & 0.01\\
\hline
\end{tabular}
\end{table}

\begin{table}[h!]
\centering
\caption{Gain, power consumption and noise figure of optimal circuit}

\begin{tabular} { |c|c|c| }
\hline
Parameter & Initial Value & Optimal Value \\
\hline
Gain (dB) & 16.35 & 13.13 \\
Power (W) & 0.01 & 0.011 \\
Noise figure (dB) & 3.56 & 2.01 \\
FoM & 459.30 & 592.67 \\
\hline
\end{tabular}
\end{table}

\begin{table}[h]
\centering
\caption{Fixed parameters list and its values
}
\begin{tabular}{|l|l|}
\hline
Parameter Name & Value \\
\hline
VDD (V) & 1.2 \\
RFin amp (V) & 0.3 \\
CMOS Length ($\mu m$)  & 0.18  \\
\hline
\end{tabular}
\end{table}

\subsection{Discussion}\label{sec4_2}
The algorithm adopted in this paper, known as CGA, has several advantages over traditional algorithms like GA when it comes to optimizing analog circuit designs. It holds multiple merits and challenges:

\subsubsection{Merits}
\textbf{1. Less Computation:} Unlike complex deep learning algorithms that require extensive training time, our algorithm utilizes minimal computation to achieve the desired specifications in circuit designs.\\
\textbf{2. Randomness \& Low Power:} Our algorithm introduces randomness to overcome issues such as overfitting and erroneous optimization points, similar to how deep learning models address these problems by incorporating momentum factors and using methods like MBGD \cite{r13}. However, deep learning techniques often involve additional parameters and higher power consumption. In contrast, our approach, which relies on randomness, offers simplicity and freedom from such challenges.\\
\textbf{3. Versatility \& Programmability:} The algorithm's simplicity allows it to be applied to various circuit blocks beyond RF circuits. Designers can easily prevent circuit components from deviating from their intended specifications, promoting flexible and tailored circuit designs. Unlike current deep learning techniques, which can become challenging when adapting to changes in input data or component ranges, our algorithm exhibits high re-programmability.

\subsubsection{Challenges}

\textbf{1. Dependency on Initial Values:} The algorithm relies on manual configuration of the ranges and initial values of all circuit components. The circuit's final optimal point can vary significantly depending on the settings, emphasizing the importance of the designer's knowledge and understanding of component ranges.\\
\textbf{2. Dependency on Randomness:} The algorithm exhibits a heuristic and prominently random nature. If the initial generation's randomly selected individuals with the best FoM are limited, the algorithm may not achieve the desired performance, even after several generations. Increasing the number of individuals in the initial generation can address this issue, but it involves a trade-off with computational resources and training time.\\
\textbf{3. Non-Convergence:} While many deep learning algorithms aim for convergence, ensuring stable algorithmic outcomes, our algorithm, with its inherently random nature, may not converge to a fixed optimal point over successive generations. While it can find circuit designs that meet the desired specifications over time, it may not reach a perfect solution.

\section{Summary}\label{sec5}
Considering the merits and challenges discussed, the algorithm used in this research is argued to offer significant convenience from the perspective of analog circuit design. Firstly, it surpasses human-optimized values within the given parameter ranges, providing enhanced performance. Secondly, it offers substantial simplicity and reduced computational requirements compared to contemporary deep learning models, which makes it more effective as the number of cascaded circuit components increases. Lastly, it promotes light, versatile, and automated optimization for various analog circuit designs, aligning with research on model compression such as pruning and bit quantization due to the increasing complexity of deep learning models \cite{r15}.

As circuit design complexity becomes more apparent, and the need for light, versatile, and efficient analog circuit optimization grows, the algorithm in this research stands as a valuable contribution to this field.

\newpage
\bibliography{sn-bibliography}

\end{document}